%% file: main.tex
\pdfoutput=1
\documentclass{article}
\usepackage{arxiv}
\usepackage{cite}
\usepackage[utf8]{inputenc} 
\usepackage{amssymb,amsfonts}
\usepackage{algorithmic}
\usepackage{graphicx}
\usepackage{textcomp}
\usepackage{xcolor}
\def\BibTeX{{\rm B\kern-.05em{\sc i\kern-.025em b}\kern-.08em
    T\kern-.1667em\lower.7ex\hbox{E}\kern-.125emX}}
\usepackage[utf8]{inputenc} 
\usepackage[T1]{fontenc}
\usepackage{url}
\usepackage{hyperref}       
\usepackage{booktabs}       
\usepackage{nicefrac}       
\usepackage{microtype}      
\usepackage{ifthen}
\usepackage[linesnumbered,ruled]{algorithm2e}
\usepackage[cmex10]{amsmath}
\usepackage{subfig}
\usepackage{wrapfig}
\usepackage[font=small]{caption}
\usepackage[colorinlistoftodos,prependcaption]{todonotes}

\title{
Generative Counterfactual Introspection for Explainable Deep Learning\\
}

\author{
Shusen Liu, Bhavya Kailkhura \\
CASC, Computation \\
Lawrence Livermore National Laboratory\\
Livermore, USA \\
\texttt{ \{liu42, kailkhura1\}@llnl.gov} 
\And
Donald Loveland, Yong Han \\
MSD, Physical and Life Science \\
Lawrence Livermore National Laboratory\\
Livermore, USA \\
\texttt{\{loveland4, han5\}@llnl.gov}
}
\begin{document}

\date{\vspace{-3ex}}

\maketitle

\begin{abstract}
In this work, we propose an introspection technique for deep neural networks that relies on a generative model to instigate salient editing of the input image for model interpretation. Such modification provides the fundamental interventional operation that allows us to obtain answers to counterfactual inquiries, i.e., what meaningful change can be made to the input image in order to alter the prediction. We demonstrate how to reveal interesting properties of the given classifiers by utilizing the proposed introspection approach on both the MNIST and the CelebA dataset. 
\end{abstract}



\input{intro.tex}
\input{relatedWorks.tex}

\input{method.tex}

\input{results.tex}
\input{discussion.tex}

\section*{Acknowledgement}
This work was performed under the auspices of the U.S. Department of Energy by Lawrence Livermore National Laboratory under Contract DE-AC52-07NA27344.

\bibliographystyle{unsrt}
\bibliography{references.bib}

\end{document}

%% file: intro.tex
\section{Introduction}
\label{sec:intro}
The recent success of deep neural networks has lead to many breakthroughs in various application domains~\cite{lecun2015deep, baldi2014searching, angermueller2016deep}.
However, these advances have also introduced increasingly complex and opaque models with decision boundaries that are extremely hard to understand.
Despite many recent developments in explainable AI, there are still enormous challenges for explaining deep neural networks. Most existing model introspection approaches~\cite{SimonyanVedaldiZisserman2013, ZeilerFergus2014, YosinskiCluneNguyen2015} focus on studying the correlation between inputs and outputs (or predictions), e.g., by identifying regions of the input image that most contributed to the final model decision. However, these methods do not consider alternative decisions or identify changes to the input which could result in different outcomes -- i.e., they are neither discriminative nor counterfactual~\cite{pearl2009causal}.
To reliably address some of the most important introspection questions, the ability to reason about causal relationships beyond correlation is necessary.

Knowing causal reasoning behind a prediction is vital in fields such as drug or material discovery~\cite{liu2017materials} where the aim is to map a known value from the output (i.e., property) space back to a set of input experimental parameters. More importantly, from a given input and output data pair, it is useful to understand how the input data could be changed to produce an output closer to their target. These necessary edits to the input data in the form of actionable knobs (implicit or explicit attribute changes) to achieve the desired results can provide a better understanding of complex decision boundaries.

A promising technique for investigating decision boundaries of a model is based on the prototype and criticism based explanations approach~\cite{kim2016examples}. In this approach, given a query sample, a prototype is defined as a quintessential data sample that best represents the class that the query sample belongs to, while a criticism is the data sample from a different target class which lay closest to the decision boundary. Explainable AI can take advantage of these relationships, as both prototype and criticism examples help build an intuitive understanding of a model and elucidate the necessary changes in the input space to achieve different responses. However, the current prototype and criticism based explanation approaches are not counterfactual in nature and cannot provide actionable feedback. Existing counterfactual explanation techniques~\cite{cxai} are limited to generating criticisms by intervening the original data space. Specifically, they generate criticisms by replacing part of the query image $I$ with specific regions of a `distractor' image $I'$ that the classifier ${C}$ predicts as class $c'$. However, making changes in the original data space (e.g., square tiles of the image)
likely will not provide actionable feedback, which is essential for many use cases, 
e.g., experimental knobs in a scientific application.
Furthermore, such changes may not be semantically meaningful and the solution space of potential explanations is restricted by the number of semantically meaningful changes in the original data space.

To overcome these limitations, in this work, we develop a \textit{generative counterfactual introspection framework} to produce inherently interpretable and actionable counterfactual visual explanations in the form of prototypes and criticisms.
The counterfactual explanation generation problem is given as follows:

\textit{Given a `query' image $I$ for which a classifier $\mathbf{C}$ predicts class $c$, a counterfactual visual explanation identifies what aspects (or attributes) of $I$ should be changed such that the classifier would output a different target class $c'$ (i.e., the criticism) or provide a more confident classification to $c$ for modified image $I'$ (i.e., the prototype).}

\begin{figure}[!tbh]
 \center
 	\includegraphics[width=0.7\textwidth]{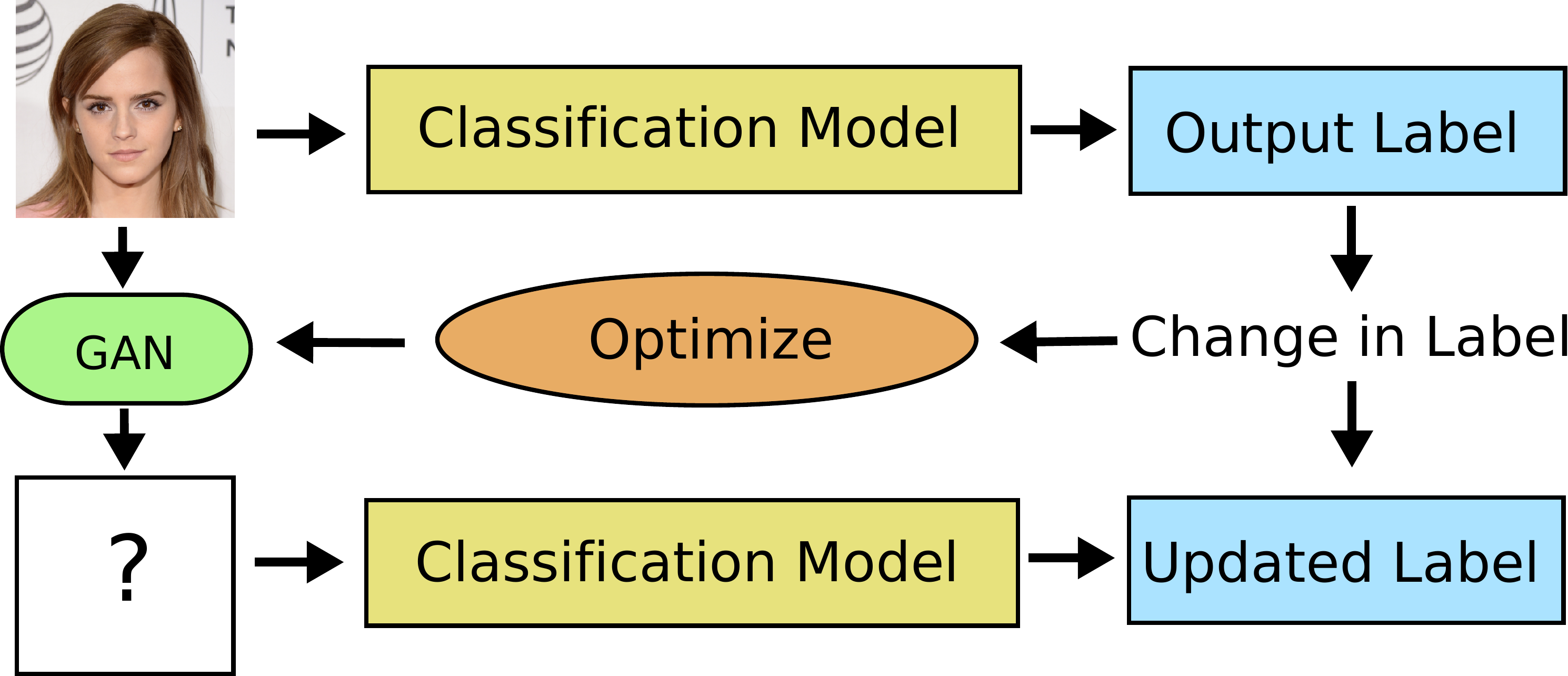}
 	\caption{The illustration of the generative counterfactual introspection concept.}
 	\label{fig:overview}
 \end{figure}

To solve this problem, we propose to employ powerful generative models along with an attribute (or actionable latent feature) editing mechanism~\cite{attGAN} to develop \textit{Generative Counterfactual Explanation}: generative and actionable counterfactual explanations generation framework (see Figure~\ref{fig:overview}). To the best of our knowledge, this is the first approach exploring the decision boundaries between classes and their relationship to the input data by providing actionable feedback and generating counterfactual prototypes and criticism based explanations.

%% file: relatedWorks.tex
\section{Related Work}
Recently, quite a few model introspection methods have been proposed to allow for interpretability of a given prediction.
Many CNN interpretation methods~\cite{SimonyanVedaldiZisserman2013, selvaraju2017grad, ZeilerFergus2014, YosinskiCluneNguyen2015, bach2015pixel}, such as GradCAM~\cite{selvaraju2017grad}, utilize backpropagation to conduct sensitivity analysis by attributing the prediction to the input domain (e.g., image pixels).
Alternatively, we can build a simpler localized model to approximate the complex nonlinear model~\cite{RibeiroSinghGuestrin2016, thiagarajan2016treeview}. 
In the LIME~\cite{RibeiroSinghGuestrin2016} work, the authors create a linear model to approximate the neural network around a specific prediction to directly attribute the prediction result into the input domain.
As proposed in~\cite{thiagarajan2016treeview}, the decision process of the neural work can also be modeled as a partition tree in the feature space. 
To understand how components of the network work, a variety of the methods have been introduced to visualization the feature (or pattern) the given neuron or layer aim to capture~\cite{nguyen2016multifaceted, erhan2009visualizing, olah2017feature} or examining the representation of the high-level concept in the latent representations~\cite{kim2017interpretability}. 

With the pressing need to obtain causal understanding of model behavior, interpretation approaches~\cite{kusner2017counterfactual,narendra2018explaining, Goyal2019, anne2018grounding} focusing on counterfactual reasoning have been proposed.
In~\cite{kusner2017counterfactual}, the counterfactual query is utilized as the fundamental tool for evaluating the fairness of the high impact social application. 
In the counterfactual visual explanation~\cite{Goyal2019} work, a patch based editing of input image is optimize in order to satisfy the intended changes in the prediction.
In the ground visual explanation~\cite{anne2018grounding} work, text based explanation are generated to provide counterfactual explanation for image classification task. 
Beside the causal interpretation methods, as demonstrated in~\cite{kim2016examples}, examining the relationship between the trained model and training dataset can also help interpret model behavior.

The safety of deep neural nets have been challenged by the existence of adversarial samples~\cite{goodfellow2014explaining, yuan2019adversarial,hogan2018universal}, in which the appearance of small but intentionally worst-case perturbations will lead to change in the prediction. 
Several specialized optimization approaches have been proposed, such as the fast gradient sign method~\cite{goodfellow2014explaining}, to resolve the optimization challenges.
Conceptually, the adversarial examples can also be considered as an answer to a counterfactual query, as it reveals a modification to the input that lead to change of the prediction. However, as the adversarial changes are imperceptible, they cannot reveal the potential bias to humans. We address this problem by utilizing generative adversarial networks (GANs)~\cite{attGAN, goodfellow2014generative} to generate modification of the input, which ensures a meaningfully edited image rather than an adversarial example.

%% file: method.tex
\section{Method}
\label{sec:method}
In order to explain a query image with respect to decision boundaries of some trained classifier on image set $\mathcal{I}$, we aim to produce counterfactual prototypes and criticisms. 
Next we formalize this problem and then present our solution.

\subsection{Minimal Change Counterfactual Example Generation}

Given a query image $I$ for which the classifier $C$ predicts class $c$, we seek to identify the key attribute changes in $I$ such that making these changes in $I$ would lead the network to either change its decision about the query to the target class (i.e., criticism) or make it more confident about the query class. We consider both of these following cases: 1) attributes are known and given for $\mathcal{I}$, or 2) attributes are unknown in which case will be learned from $\mathcal{I}$. Furthermore, these attributes are expected to be actionable, i.e., we should be able to change these attributes and generate corresponding changes in the query image. To enable this, we employ a powerful generative machine learning model called ``generative adversarial network (GAN)''~\cite{goodfellow2014generative}. GANs transform vectors of generated noise (or latent factors) into synthetic samples resembling data gathered in the training set. GANs (and corresponding latent space) are learned in an adversarial manner, i.e., a concept taken from the game theory which assumes two
competing networks, a discriminator $D$ (differentiating real vs. synthetic samples) and a generator $G$ (learning to produce realistic synthetic samples by transforming latent factors). This adversarial learning is shown to learn salient attributes of the data in an unsupervised manner which can later be manipulated using the generator $G$. GANs can also be used for simultaneously generating and manipulating the images with known and desired attributes~\cite{attGAN}. We use both of these formulations in our framework depending on whether actionable attributes are known or unknown, where the latter uses the latent representations as our attributes. 

Generative editing models are denoted as ${G}(I;A)$ or ${G}(I; L_{o})$ depending on whether actionable attributes are known or unknown respectively. The goal is to manipulate single or multiple attributes ${A}=\{a_1,\cdots,a_N\}$ of an image $I$, i.e., to generate a new image $I^*$ with desired attributes $\{a_1^*,\cdots,a_M^*\}$ while preserving other details $\{a_{M+1},\cdots,a_N\}$, or to manipulate a latent vector $L_{o}$ in a similar fashion. 
Given these generative editing mechanism, we formulate minimal change counterfactual explanation generation problem given image $I$, image attribute $A$, and a target attribute vector $A'$, where $L_{o}$ and $L_{o}'$ can be used in place of $A$ and $A'$, as follows:

\begin{equation}
\label{opt}
\begin{aligned}
\min_{A'} \quad & \|I-I(A')\|_p\\
\textrm{s.t.} \quad & c'=C(I(A'))\\
  & I(A') = G(I;A')    \\
\end{aligned}
\end{equation}
where $p= 1$ and $c'$ is the target criticism class.
When the goal is to generate prototypes, we set $c'=c$ as the original class label of the query image and formulate an alternating loss function to promote solution which maximize class confidence instead of having a trivial solution, i.e., $A'=A$. 

\subsection{Approximate Solution}
Most deep neural network based models make formulation \eqref{opt} non-linear and non-convex, making it hard to find a closed-form solution. Thus, we formulate a relaxed version of this optimization problem which can be solved efficiently using gradient descent algorithms. 
The proposed approach relaxes the optimization problem \ref{opt} as follows:
\begin{equation}
\label{opt1}
\begin{aligned}
\min_{A'} \quad & \lambda \cdot loss_{C,c'}(I(A'))+  \|I- I(A')\|_p\\
\end{aligned}
\end{equation}
where loss $loss_{C,c'}$ is cross-entropy loss for predicting image $I(A')$ to label $c'$ using classifier $C$. Note that both classifier $C$ and generator $G$ are differentiable. The gradient of the objective function is computed by back-propagation, and the minimal change counterfactual example generation problem is solved using gradient descent.
Furthermore, to generate an explanation with minimum change $\delta=\|I- I(A')\|_p$, one can repeatedly solve this optimization problem using gradient descent, continually updating $\lambda$ using bisection search or any other method for one-dimensional optimization.

%% file: results.tex
\section{Experiments}
Here we demonstrate the effectiveness of the proposed counterfactual explanation generation approach on two datasets (one with known attributes and another one with unknown).
The proposed method outputs modified images to satisfy counterfactual queries along with actionable attribute values to achieve these results, in turn, providing a comprehensive understanding of decision boundaries of the classifier $C$.

\subsection{MNIST dataset}
In this experiment, we consider the problem of
classifying a given image of a handwritten digit into one of 10 classes (0 to 9). We use the MNIST dataset~\cite{mnist_bib} which contains 60,000 training and 10,000 test images of handwritten digits. The classifier~\cite{lenet} is trained on MNIST training set and achieves $99.10\%$ accuracy on the test set. We utilize a pretrained DCGAN architecture~\cite{dcgan} (with a 10D latent space) as our image generator.
Given 10D latent vector ($L_{o}$), the generator produces a digit image.
The proposed optimization method will update the $L_{o}$ to generate meaningful modification of the image that answers the counterfactual query.

\begin{figure}[!tbh]
\center
	\includegraphics[width=0.6\textwidth]{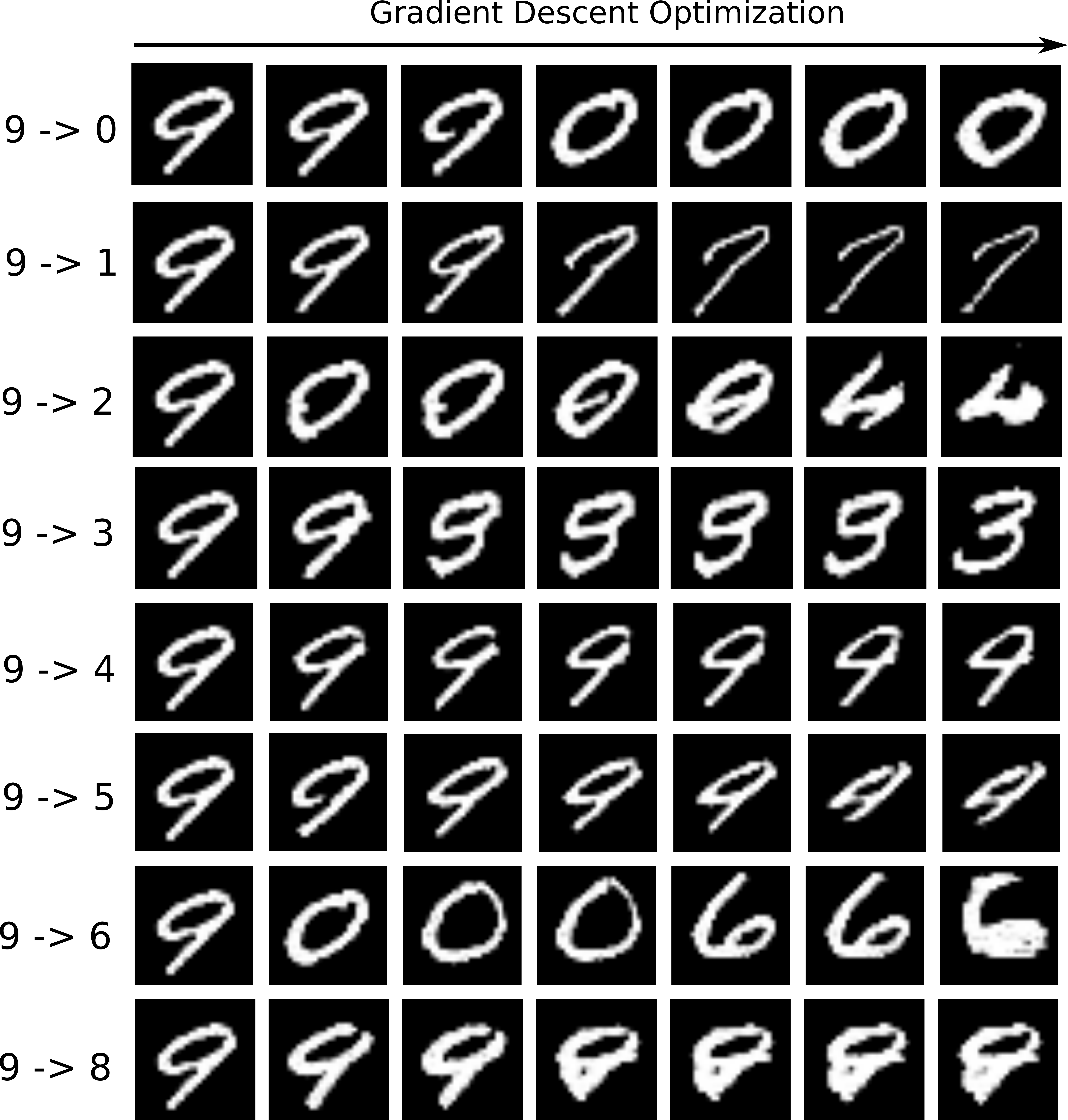}
	\caption{Finding criticism of the digit $9$ class.}
	\label{fig:MNIST_criticism}
\end{figure}

\begin{figure}[!tbh]
\center
	\includegraphics[width=0.6\textwidth]{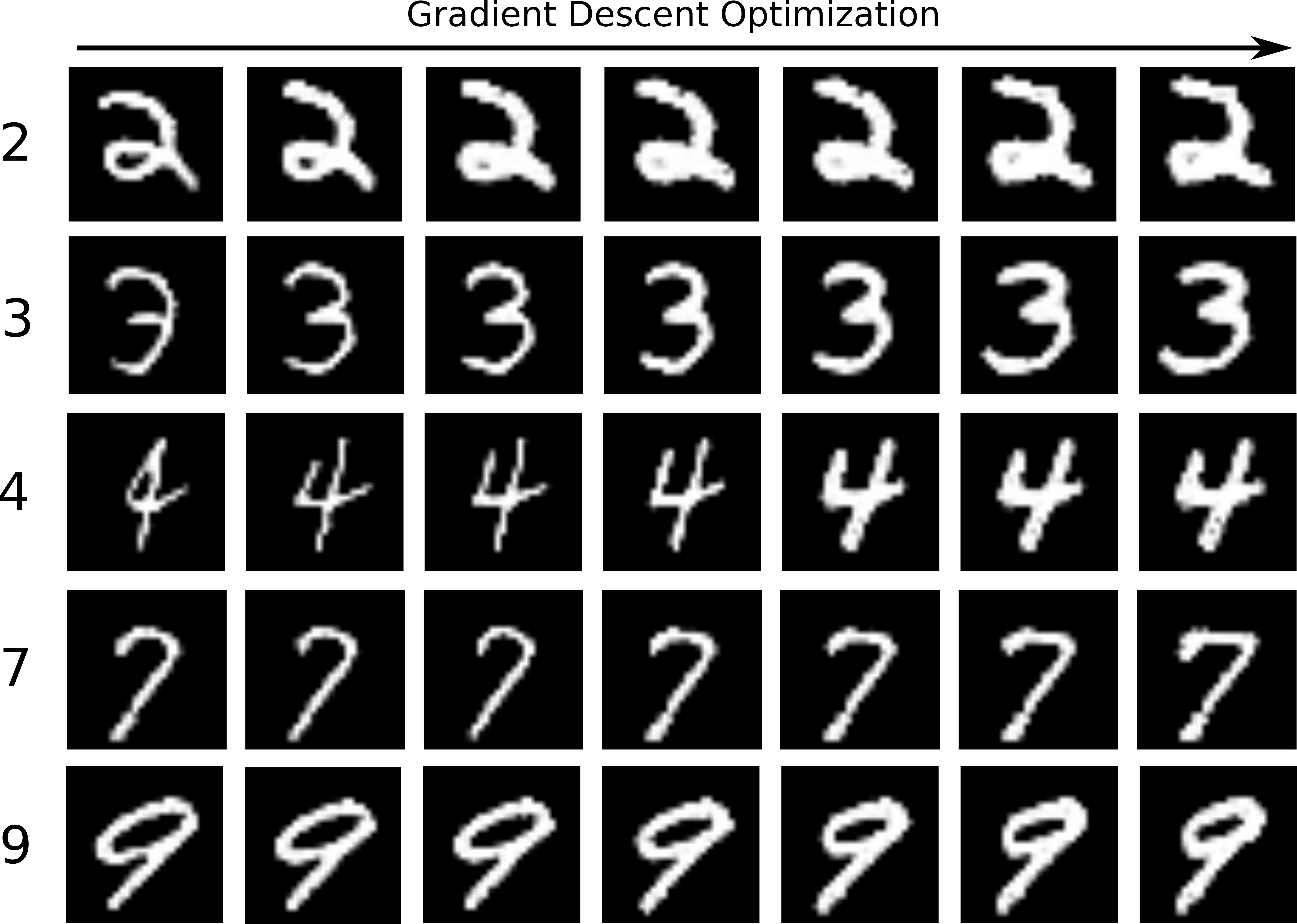}
	\caption{Finding prototypes of different digits.}
	\label{fig:MNIST_prototype}
\end{figure}

As shown in Figure~\ref{fig:MNIST_criticism}, we illustrate meaningful changes to the image of digit $9$ to alter its prediction.
We start from the same image in each row and illustrate the optimization path from the original image to the images that altered the classified label to a predefined target label.
Compared to a direction optimization in the image space~\cite{goodfellow2014explaining} that leads to an adversarial example, the utilization of a GAN guarantees that we end up exploring the ``manifold'' of all possible meaningful images.
As a result, these edits provide us with valuable insights regarding classifier decision boundary, i.e., what are the boundary image patterns between different classes of digits, and what kind of changes are most likely to alter the prediction.
Interestingly, we see that for certain target labels ($9$->$2$, $9$->$6$), the image first change to a different digit (in this case $0$) before morphing into the target digits.
Alternatively, as shown in Figure~\ref{fig:MNIST_prototype}, we also utilize a similar optimization to find the \emph{prototype} for each digit by``walking'' toward the center of the class on the digit image manifold. We can see the starting digits morphed into a more ``regular'' handwriting style, which are easier for human to recognize.
These observations not only help in revealing the inherent structure of the digit image manifold but also indicate the preference of the classifier regarding similarity between digits.

%

\begin{figure}[!tbh]
\center
	\includegraphics[width=0.7\textwidth]{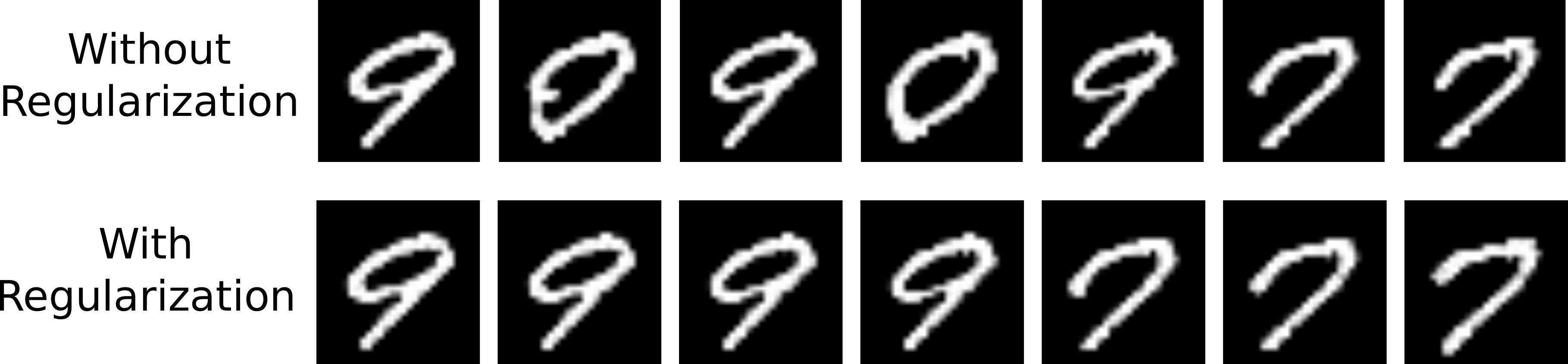}
	\caption{The effect of regularization on the optimization path for finding criticism of 9 in the direction of 7.}
	\label{fig:MNIST_reg}
	\vspace{-4mm}
\end{figure}

For better interpretability, it is desirable to make sure the modification of the image is relatively small and consistent. As discussed in Section~\ref{sec:method}, we include a regularization term that measure the distance between the original image and the edited ones. In Figure~\ref{fig:MNIST_reg}, we can see that this regularization ensure the optimization process is smooth and the modifications to the images are keep to the minimal.

\subsection{CelebA dataset}
\begin{figure}[!tbh]
 \center
 	\includegraphics[width=0.55\textwidth]{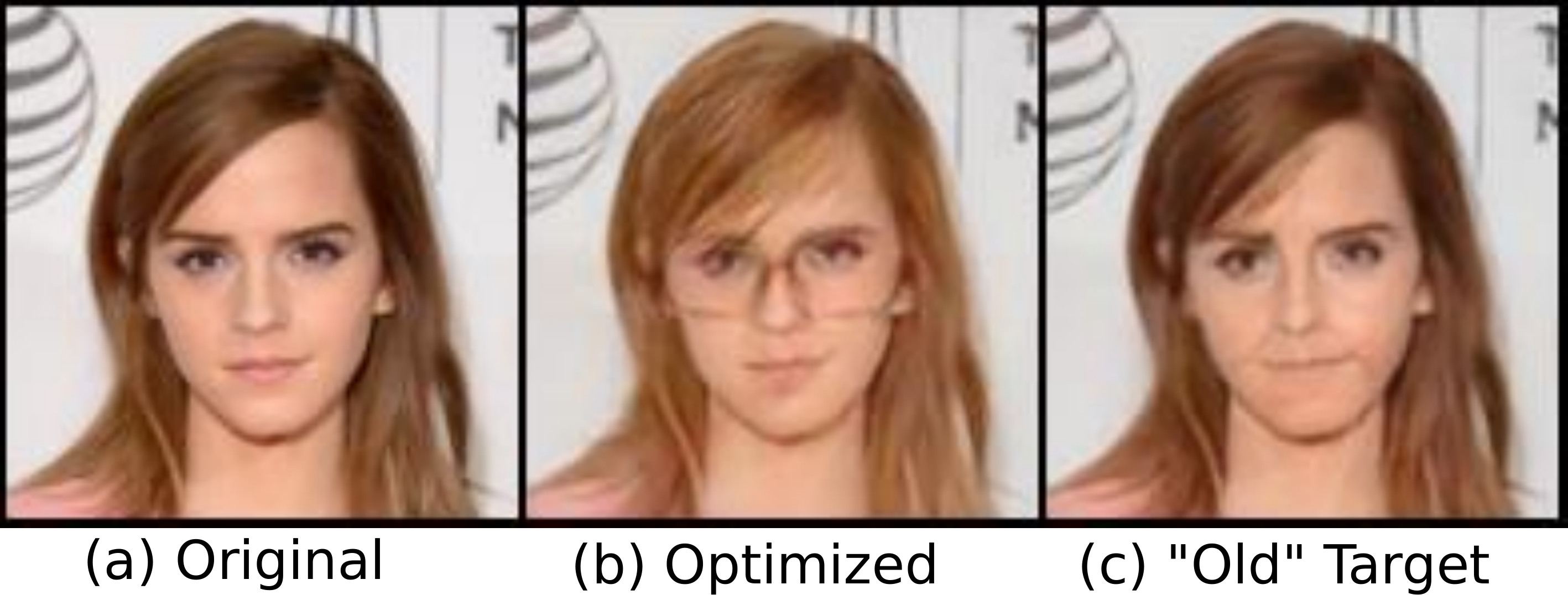}
 	\caption{Illustrate different editing scheme for the input image. The original image is shown in (a). In (c), we show the edited image that predicted as "old" by altering the "old/young" attribute of the AttGAN. In (b), we show the modification of the same image driven by the preference of the classier (without modifying the "old/young" attribute). 
 }
 \label{fig:emma}
 \end{figure}

In several case, attributes are known explicitly, thus, optimization can be carried out in the attribute space ${A}$ that a generator is conditioned on ($I'={G}(I; {A})$), where explicitly defined physical attributes can provide actionable feedback.
We use CelebFaces Attributes dataset (CelebA)~\cite{liu2015deep}, which is a large-scale face attributes dataset with more than 200K celebrity images, each with explicit attribute annotations. We consider a classification problem of classifying a celebrity face image in CelebA dataset into young or old. The classifier~\cite{attGAN} is trained on CelebA dataset and achieves average accuracy of $90.89\%$ on CelebA testing set. Next, we use the AttGAN~\cite{attGAN} as our generative editing method to generate modification to the query face image.
The AttGAN can make edits to the original query image $I$ based on additional attributes (e.g., hair color, glass, bang, bald).
See Figure~\ref{fig:emma}(a)(c), we can make the image in (a) looks older by setting the old/young attribute when generate the new image (c), where features such as wrinkle are added to make the subject appears to appear older.
Such a generator allows us to only change a given person's superficial appearance without alter facial features and identify, which also allow us to obtain more meaningful counterfactual explanation for probing the behavior of the given classifier.
In the following experiments, we focus on exploring the behavior of a classifier trained for predicting whether a person is young or old based on the given image.

Since the AttGAN generator has a young/old input attribute, a direct optimization in the entire attribute space will likely lead to the degenerate case, in which the young/old attribute is used to edit the image (to make it appears older for the classifier). Therefore, in our experiment, we fixed the young/old attribute to the original label and only make changes to rest of the attributes (12 in total). In other words, we ask what kind of attributes changes (beside the young/old attribute) will make a given image appear older or younger for the given classifier.

\begin{figure*}
\center
\includegraphics[width=0.98\textwidth]{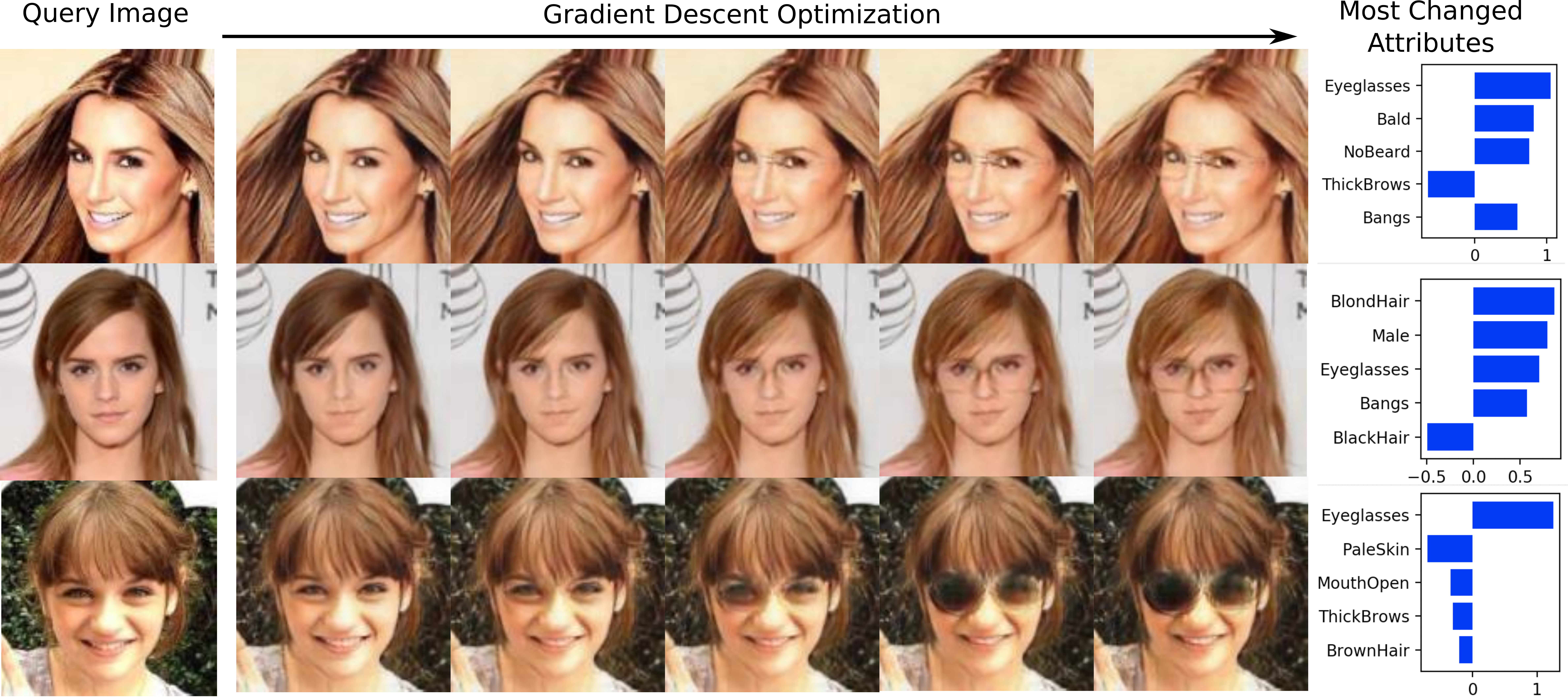}
\caption{
Illustrate attributes changes (beside the young/old attribute) that will make the images appear older for the given classifier.
The left most column shows the original image.
The right most column shows the top five most changed attributes and their relatively changes.
}
\vspace{-4mm}
\label{fig:celeba_young_old}
\end{figure*}

\begin{figure}[!tbh]
\center
 \includegraphics[width=0.6\textwidth]{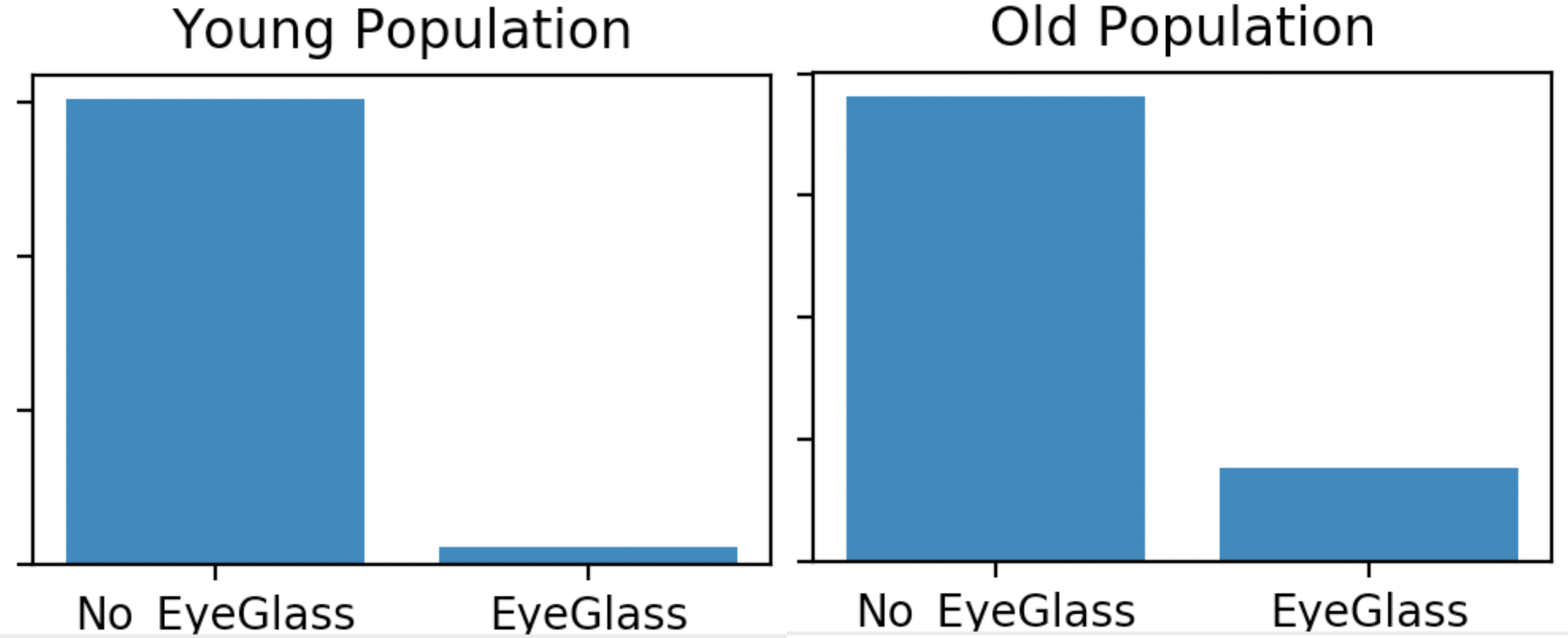}
 \caption{The potential bias in the CelebA dataset. The percentage of people having eye glass is much higher in the population labeled as ``old''.}
 \label{fig:celeba_eyeGlass}
\end{figure}

In Figure~\ref{fig:celeba_young_old}, we have three female celebrity faces (query images) which are classified as ``young''. Here, we show the optimization path that eventually leads to an ``old'' classification. The right most column shows the top five most changed attributes and their relatively changes. This result is particularly interesting as all three examples show eyeglasses in the modified images that result in an ``old'' classification.
One possible explanation for such an observation is that the classifier learns these patterns from the training data. To investigate this hypothesis, we explore the distributions of attributes across the training data.
As shown in Figure~\ref{fig:celeba_eyeGlass}, we can see a clear difference regarding eye glass frequency between the young and old population.
This result demonstrate that counterfactual query can be an very powerful tool to reveal unexpected behaviors of classifiers and highlight the potential bias in the training data.

\begin{figure*}
\center
	\includegraphics[width=0.99\textwidth]{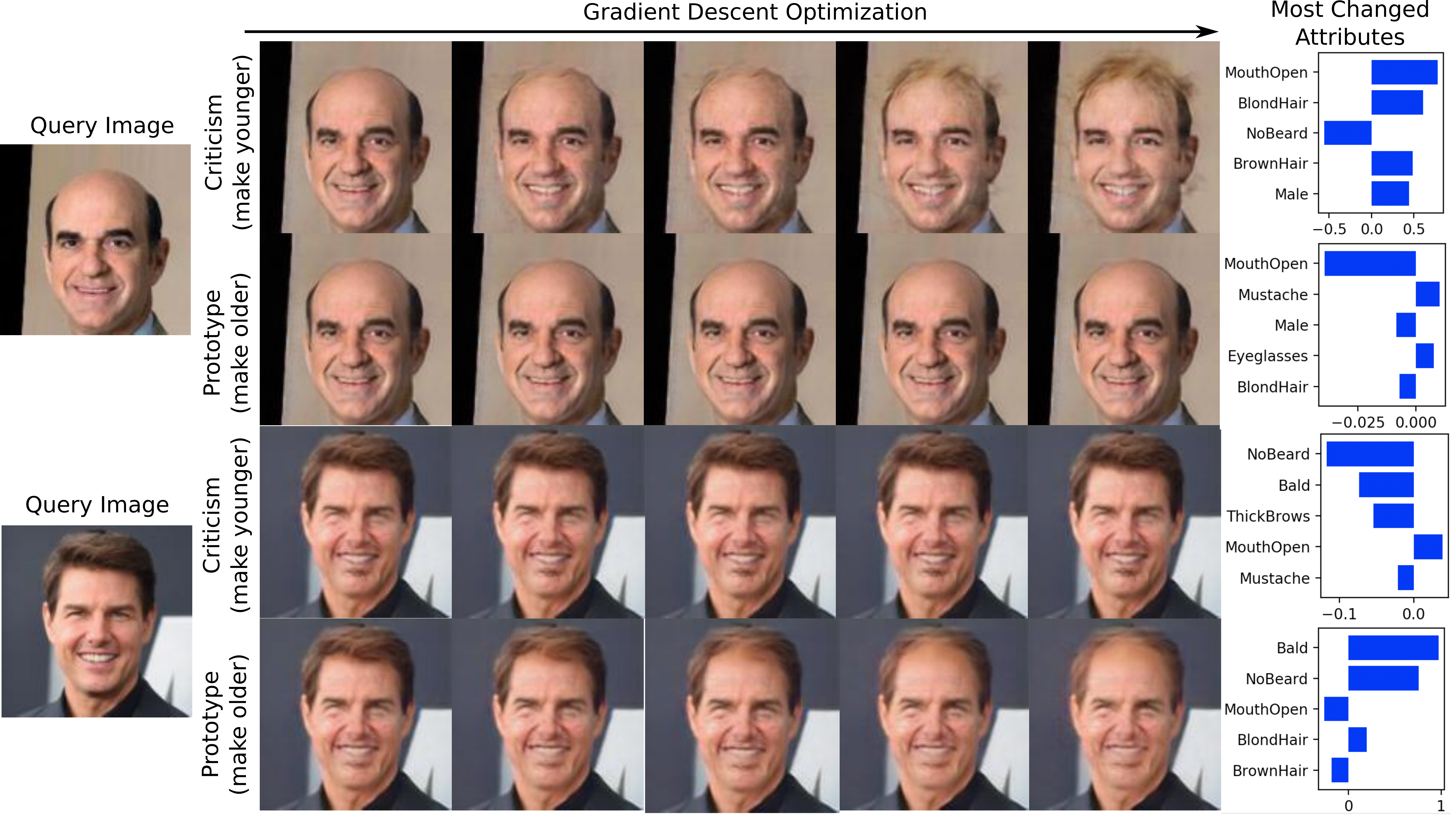}
	\vspace{-1mm}
	\caption{Prototype and criticism for the images with ground truth label ``old''. The left most column shows the original image. The right most column shows the top five most changed attributes and their relatively changes.
	}
	\vspace{-4mm}
	\label{fig:celeba_old_young}
\end{figure*}

To further illustrate how counterfactual examples help explain the behavior of the classifier, in Figure~\ref{fig:celeba_old_young}, we investigate the prototype and criticism examples for two male celebrity faces that both have a ground truth label ``old''.
When searching for the prototypes (i.e., making them older), we see a minimal changes for the first face (second row) while observe significant change for the second face (fourth row). 
This distinction indicates that the first person seems to have a prototypical look for the ``old'' class, whereas the second person does not.
For the criticisms (row one and three), the opposite holds true, which indicates the image of the second person is an outlier for ``old'' samples, and is closer to a typical ``young'' image.
Finally, the right most column provides ``actionable insights'' to achieve these changes. The top five most changed attributes are reasonable with hair features being most important factors in discriminating the age group.







%% file: discussion.tex
\section{Discussion and Future Work}
In this work, we present preliminary results on utilizing generative models to obtain counterfactual explanations for a given classifier. Despite the simplicity of the optimization, we demonstrate that the effectiveness of the proposed approach for revealing insights regarding the behavior of deep neural network models. 
For future directions, we plan to explore the potential application of such interpretation method for scientific application, where explainability are essential for model validation and domain discovery.  

%% file: main.bbl
\begin{thebibliography}{10}

\bibitem{lecun2015deep}
Yann LeCun, Yoshua Bengio, and Geoffrey Hinton.
\newblock Deep learning.
\newblock {\em nature}, 521(7553):436, 2015.

\bibitem{baldi2014searching}
Pierre Baldi, Peter Sadowski, and Daniel Whiteson.
\newblock Searching for exotic particles in high-energy physics with deep
  learning.
\newblock {\em Nature communications}, 5:4308, 2014.

\bibitem{angermueller2016deep}
Christof Angermueller, Tanel P{\"a}rnamaa, Leopold Parts, and Oliver Stegle.
\newblock Deep learning for computational biology.
\newblock {\em Molecular systems biology}, 12(7):878, 2016.

\bibitem{SimonyanVedaldiZisserman2013}
Karen Simonyan, Andrea Vedaldi, and Andrew Zisserman.
\newblock Deep inside convolutional networks: Visualising image classification
  models and saliency maps.
\newblock {\em arXiv preprint arXiv:1312.6034}, 2013.

\bibitem{ZeilerFergus2014}
Matthew~D Zeiler and Rob Fergus.
\newblock Visualizing and understanding convolutional networks.
\newblock In {\em European conference on computer vision}, pages 818--833.
  Springer, 2014.

\bibitem{YosinskiCluneNguyen2015}
Jason Yosinski, Jeff Clune, Anh Nguyen, Thomas Fuchs, and Hod Lipson.
\newblock Understanding neural networks through deep visualization.
\newblock {\em arXiv preprint arXiv:1506.06579}, 2015.

\bibitem{pearl2009causal}
Judea Pearl et~al.
\newblock Causal inference in statistics: An overview.
\newblock {\em Statistics surveys}, 3:96--146, 2009.

\bibitem{liu2017materials}
Yue Liu, Tianlu Zhao, Wangwei Ju, and Siqi Shi.
\newblock Materials discovery and design using machine learning.
\newblock {\em Journal of Materiomics}, 3(3):159--177, 2017.

\bibitem{kim2016examples}
Been Kim, Rajiv Khanna, and Oluwasanmi~O Koyejo.
\newblock Examples are not enough, learn to criticize! criticism for
  interpretability.
\newblock In {\em Advances in Neural Information Processing Systems}, pages
  2280--2288, 2016.

\bibitem{cxai}
Yash Goyal, Ziyan Wu, Jan Ernst, Dhruv Batra, Devi Parikh, and Stefan Lee.
\newblock Counterfactual visual explanations.
\newblock {\em CoRR}, abs/1904.07451, 2019.

\bibitem{attGAN}
Zhenliang He, Wangmeng Zuo, Meina Kan, Shiguang Shan, and Xilin Chen.
\newblock Attgan: Facial attribute editing by only changing what you want.
\newblock {\em IEEE Transactions on Image Processing}, 2019.

\bibitem{selvaraju2017grad}
Ramprasaath~R Selvaraju, Michael Cogswell, Abhishek Das, Ramakrishna Vedantam,
  Devi Parikh, and Dhruv Batra.
\newblock Grad-cam: Visual explanations from deep networks via gradient-based
  localization.
\newblock In {\em Proceedings of the IEEE International Conference on Computer
  Vision}, pages 618--626, 2017.

\bibitem{bach2015pixel}
Sebastian Bach, Alexander Binder, Gr{\'e}goire Montavon, Frederick Klauschen,
  Klaus-Robert M{\"u}ller, and Wojciech Samek.
\newblock On pixel-wise explanations for non-linear classifier decisions by
  layer-wise relevance propagation.
\newblock {\em PloS one}, 10(7):e0130140, 2015.

\bibitem{RibeiroSinghGuestrin2016}
Marco~Tulio Ribeiro, Sameer Singh, and Carlos Guestrin.
\newblock Why should i trust you?: Explaining the predictions of any
  classifier.
\newblock In {\em Proceedings of the 22nd ACM SIGKDD International Conference
  on Knowledge Discovery and Data Mining}, pages 1135--1144. ACM, 2016.

\bibitem{thiagarajan2016treeview}
Jayaraman~J Thiagarajan, Bhavya Kailkhura, Prasanna Sattigeri, and
  Karthikeyan~Natesan Ramamurthy.
\newblock Treeview: Peeking into deep neural networks via feature-space
  partitioning.
\newblock {\em arXiv preprint arXiv:1611.07429}, 2016.

\bibitem{nguyen2016multifaceted}
Anh Nguyen, Jason Yosinski, and Jeff Clune.
\newblock Multifaceted feature visualization: Uncovering the different types of
  features learned by each neuron in deep neural networks.
\newblock {\em arXiv preprint arXiv:1602.03616}, 2016.

\bibitem{erhan2009visualizing}
Dumitru Erhan, Yoshua Bengio, Aaron Courville, and Pascal Vincent.
\newblock Visualizing higher-layer features of a deep network.
\newblock {\em University of Montreal}, 1341(3):1, 2009.

\bibitem{olah2017feature}
Chris Olah, Alexander Mordvintsev, and Ludwig Schubert.
\newblock Feature visualization.
\newblock {\em Distill}, 2(11):e7, 2017.

\bibitem{kim2017interpretability}
Been Kim, Martin Wattenberg, Justin Gilmer, Carrie Cai, James Wexler, Fernanda
  Viegas, and Rory Sayres.
\newblock Interpretability beyond feature attribution: Quantitative testing
  with concept activation vectors (tcav).
\newblock {\em arXiv preprint arXiv:1711.11279}, 2017.

\bibitem{kusner2017counterfactual}
Matt~J Kusner, Joshua Loftus, Chris Russell, and Ricardo Silva.
\newblock Counterfactual fairness.
\newblock In {\em Advances in Neural Information Processing Systems}, pages
  4066--4076, 2017.

\bibitem{narendra2018explaining}
Tanmayee Narendra, Anush Sankaran, Deepak Vijaykeerthy, and Senthil Mani.
\newblock Explaining deep learning models using causal inference.
\newblock {\em arXiv preprint arXiv:1811.04376}, 2018.

\bibitem{Goyal2019}
Jan Ernst Dhruv Batra Devi Parikh Stefan~Lee Yash~Goyal, Ziyan~Wu.
\newblock Counterfactual visual explanations.
\newblock In {\em ICML}, pages 264--279, 2019.

\bibitem{anne2018grounding}
Lisa Anne~Hendricks, Ronghang Hu, Trevor Darrell, and Zeynep Akata.
\newblock Grounding visual explanations.
\newblock In {\em Proceedings of the European Conference on Computer Vision
  (ECCV)}, pages 264--279, 2018.

\bibitem{goodfellow2014explaining}
Ian~J Goodfellow, Jonathon Shlens, and Christian Szegedy.
\newblock Explaining and harnessing adversarial examples.
\newblock {\em arXiv preprint arXiv:1412.6572}, 2014.

\bibitem{yuan2019adversarial}
Xiaoyong Yuan, Pan He, Qile Zhu, and Xiaolin Li.
\newblock Adversarial examples: Attacks and defenses for deep learning.
\newblock {\em IEEE transactions on neural networks and learning systems},
  2019.

\bibitem{hogan2018universal}
Thomas~A Hogan and Bhavya Kailkhura.
\newblock Universal hard-label black-box perturbations: Breaking
  security-through-obscurity defenses.
\newblock {\em arXiv preprint arXiv:1811.03733}, 2018.

\bibitem{goodfellow2014generative}
Ian Goodfellow, Jean Pouget-Abadie, Mehdi Mirza, Bing Xu, David Warde-Farley,
  Sherjil Ozair, Aaron Courville, and Yoshua Bengio.
\newblock Generative adversarial nets.
\newblock In {\em Advances in neural information processing systems}, pages
  2672--2680, 2014.

\bibitem{mnist_bib}
Yann LeCun and Corinna Cortes.
\newblock {MNIST} handwritten digit database.
\newblock 2010.

\bibitem{lenet}
Yann LeCun, L{\'e}on Bottou, Yoshua Bengio, Patrick Haffner, et~al.
\newblock Gradient-based learning applied to document recognition.
\newblock {\em Proceedings of the IEEE}, 86(11):2278--2324, 1998.

\bibitem{dcgan}
Alec Radford, Luke Metz, and Soumith Chintala.
\newblock Unsupervised representation learning with deep convolutional
  generative adversarial networks.
\newblock {\em arXiv preprint arXiv:1511.06434}, 2015.

\bibitem{liu2015deep}
Ziwei Liu, Ping Luo, Xiaogang Wang, and Xiaoou Tang.
\newblock Deep learning face attributes in the wild.
\newblock In {\em Proceedings of the IEEE international conference on computer
  vision}, pages 3730--3738, 2015.

\end{thebibliography}
